\documentclass[12pt, a4paper]{article}
\usepackage[utf8]{inputenc}
\usepackage{amsmath, amssymb, amsthm}
\usepackage{graphicx}
\usepackage{tikz}
\usetikzlibrary{positioning,arrows.meta,calc}
\usepackage{geometry}
\usepackage{setspace}
\usepackage{algorithm}
\usepackage{algorithmic}
\usepackage{hyperref}
\newtheorem{definition}{Definition}[section]
\newtheorem{theorem}{Theorem}[section]
\newtheorem{lemma}[theorem]{Lemma}
\newtheorem{assumption}[theorem]{Assumption}
\theoremstyle{remark}
\newtheorem{remark}[theorem]{Remark}
\theoremstyle{plain}

\newcommand{\R}{\mathbb{R}}
\DeclareMathOperator*{\argmin}{arg\,min}

\title{\large\textbf{Fiber Bundle Networks: Interpretable Machine Learning via\\ Riemannian Prototype Optimization}}
\author{Skyler D. Liu\thanks{Corresponding author: liudong@mit.edu}\textsuperscript{\phantom{*}1} \\[0.6em]
{\footnotesize $^{1}$Massachusetts Institute of Technology, Cambridge MA, 02139}}
\date{}

\begin{document}

\maketitle

\begin{abstract}
We propose Fiber Bundle Networks (FiberNet), a machine learning framework grounded in differential geometry. Unlike conventional deep neural networks, which rely on black-box function fitting, we reformulate classification as interpretable geometric optimization on fiber bundles, where categories form the base space and wavelet-transformed features lie in the fibers above each category. FiberNet jointly learns a Riemannian metric that highlights informative frequency components and a set of class prototypes, obtained by minimizing an energy that balances data fidelity, semantic smoothness, and class separation. A test sample is assigned to its nearest prototype, so each class corresponds to an explicit Voronoi region in feature space. We prove that the proposed alternating optimization monotonically decreases the energy and that the energy sequence converges. Since FiberNet requires only $Nd + d$ parameters for $N$ classes and $d$ features, this suggests that fiber bundle geometry offers a principled route to models that are both interpretable and parameter-efficient, two properties that are difficult to obtain simultaneously in conventional deep learning.

\end{abstract}

\section{Introduction}

Consider the task of distinguishing the sound of ocean waves from the sound of a chair scraping against the floor. Ocean waves produce low-frequency periodic fluctuations (20--500~Hz) that evolve slowly over several seconds, whereas chair scraping produces high-frequency aperiodic noise (2000--8000~Hz) with abrupt onsets. The two sounds thus differ fundamentally in their spectral characteristics. How can an algorithm automatically learn that ``low frequencies are more important for ocean waves'' while ``high frequencies are more important for scraping sounds''? How can we leverage the semantic knowledge that ``ocean waves are similar to rain sounds but far from scraping sounds''?

Deep neural networks learn such relationships implicitly through stacked nonlinear layers \cite{lecun2015deep}; although powerful, they offer little insight into what has been learned or why it works \cite{zhang2021survey,lipton2018mythos}. In contrast, FiberNet models these relationships explicitly within an interpretable geometric framework: a learnable Riemannian metric adaptively weights the importance of different frequency-band features, and a variational energy functional injects semantic structure among categories.

FiberNet is applicable to a range of classification tasks, including audio, image, text, and genomic sequence classification. Each data modality calls for an appropriate feature extractor: wavelet transforms capture the time--frequency structure of audio, multi-scale filter banks extract texture features from images, word embeddings capture semantic relationships in text, and oligonucleotide frequency statistics summarize patterns in gene sequences. In this paper, we use audio classification as a running example to present the FiberNet mechanism.

The dominant approach to audio classification relies on deep neural networks, for example convolutional networks that learn representations directly from raw waveforms \cite{abdoli2019end} or from log-mel spectrograms \cite{kong2020panns}. Despite their success, three challenges remain. First, large models have substantial parameter counts (approximately 11M for ResNet-18 \cite{he2016deep} and 80M for the CNN14 model of PANNs \cite{kong2020panns}), which hinders deployment on resource-constrained edge devices such as smart speakers and hearing aids. Second, these models are largely black boxes: they offer limited insight into why a sound is classified as ``ocean waves'' rather than ``rain'', which is a major obstacle in high-stakes applications such as medical acoustic diagnosis. Third, the standard cross-entropy loss ignores semantic relationships between categories. For instance, ``ocean waves'' and ``rain'' are both natural sounds and should be closer to each other than to ``metal striking'', yet cross-entropy penalizes all misclassifications equally.

To address these challenges, FiberNet makes the learning process explicit through fiber bundles. Inspired by recent uses of fiber bundles in machine learning \cite{courts2021bundle}, its core idea is to cast classification as a geometric optimization problem on a fiber bundle whose base space is the set of categories and whose fiber over each category contains the audio features of that category. This formulation explicitly encodes the two-level ``category--feature'' structure and allows us to employ geometric tools (sections and Riemannian metrics) together with variational optimization.

FiberNet consists of three core components. The first is a learnable diagonal Riemannian metric $G = \operatorname{diag}(a_1, \ldots, a_d)$, whose weights $a_i > 0$ adaptively reflect the discriminative power of each wavelet feature. The Euclidean distance treats all frequency bands equally, whereas in audio classification the relevance of different bands varies considerably: ocean wave sounds depend mainly on low-frequency components (20--500~Hz), while whistles depend on high-frequency components (1000--4000~Hz). Learning the metric from data \cite{scarvelis2023riemannian} enables the algorithm to identify informative frequency bands automatically and to suppress noisy or irrelevant dimensions.

The second component is variational optimization through energy minimization. Each category prototype serves as a representative feature vector for its class and is defined as the value of a section of the fiber bundle at that category, where the section maps the category space to the feature space. The prototypes are optimized by minimizing an energy functional with three terms: an attachment energy that draws each prototype toward the training samples of its category; a tension energy that encourages prototypes of semantically similar categories (e.g., ``ocean waves'' and ``rain'') to be close; and a repulsion energy that keeps prototypes of different categories (e.g., ``ocean waves'' and ``metal striking'') sufficiently separated.

The third component is a classification rule based on Voronoi partitioning. Training yields $N$ category prototypes, where $N$ is the number of categories. At inference time, a sample is assigned to the category whose prototype is nearest under the learned Riemannian distance. Equivalently, the prototypes partition the feature space into $N$ Voronoi cells, and classification amounts to determining the cell in which a new sample lies.

Our main contributions are as follows.
\begin{itemize}
\item We propose Fiber Bundle Networks, a machine learning approach that, to the best of our knowledge, is the first to systematically apply fiber bundle theory from differential geometry to classification.
\item The learned Riemannian metric weights reveal the importance of individual features in each frequency band, and energy minimization yields prototype configurations that encode semantic relationships among categories.
\item We develop a variational algorithm that jointly optimizes the metric and the prototypes, and we prove that the resulting energy sequence decreases monotonically and converges.
\end{itemize}

\section{Related Work}

Fiber bundles, a core concept in differential geometry, have only recently been introduced into machine learning and data analysis, and related work remains scarce. Courts and Kvinge~\cite{courts2021bundle} proposed Bundle Networks for exploring the fiber structure of many-to-one mappings. Their key idea is to use local trivializations to decompose the space into a product of base space and fiber, which naturally models a generative process: given an output $y$, generate the set of inputs $x$ satisfying $f(x) = y$ (i.e., the fiber over $y$). This work is conceptually dual to ours. Bundle Networks address the generative problem (recovering the diversity of inputs from an output), whereas FiberNet addresses the inverse, discriminative problem (mapping inputs to outputs). Our section selects a representative point in the fiber of each category, whereas Bundle Networks generate the entire fiber over a given point. Both approaches exploit the local structure of fiber bundles but pursue complementary objectives: Bundle Networks aim to capture generative diversity, whereas FiberNet aims to separate categories.

Geometric structures have also been exploited in other areas of machine learning, including Riemannian frameworks for statistics on manifold-valued data \cite{pennec2006riemannian} and manifold learning \cite{tenenbaum2000global}. The FibeRed algorithm of Scoccola and Perea~\cite{scoccola2022fibered} uses vector bundles for dimensionality reduction from the perspective of topological data analysis, modeling data as vector bundles so that the base space captures large-scale topology while fibers capture local geometry. Like FibeRed, our approach uses the hierarchical structure of a fiber bundle to model complex data, with the base space carrying global information and the fibers carrying local detail. The difference is that FibeRed's base space is the topological structure of the data (discovered through topological inference), while FiberNet's base space is the category space, and classification is achieved by learning a section (a map from categories to features) and a metric (the geometry of the feature space).

FiberNet is also related to prototype-based classification and metric learning. The nearest-prototype rule is a compressed form of nearest-neighbor classification \cite{cover1967nearest}, in which each class is represented by a single point rather than by all of its training samples. Weighting feature dimensions by a learned positive-definite matrix generalizes the Mahalanobis distance \cite{mahalanobis1936generalized}, and learning Riemannian metrics from data has recently been studied from an optimal transport perspective \cite{scarvelis2023riemannian}. FiberNet differs from these approaches in that the prototypes and the metric are optimized jointly under a single variational energy that additionally encodes semantic relations between categories.

\section{Mathematical Framework of FiberNet}

\subsection{Fiber Bundle Structure}\label{sec:bundle}

We present FiberNet using audio classification as a running example. A fiber bundle is a geometric structure that naturally represents categories, features, and the relationships between them. Let $N$ denote the number of categories, $C = \{c_1, c_2, \ldots, c_N\}$ the category set, and $d$ the feature dimension.

\begin{definition}[Classification Fiber Bundle]
The fiber bundle for an $N$-class classification problem is a triple $(\mathcal{E}, C, \pi)$, where:
\begin{itemize}
\item the \emph{base space} is the category set $C$, endowed with the discrete topology;
\item the \emph{total space} $\mathcal{E} = C \times \R^d$ is the set of all category--feature pairs;
\item the \emph{projection} $\pi: \mathcal{E} \to C$, $\pi(c, v) = c$, extracts the category label;
\item the \emph{fiber} over $c$ is $F_c = \pi^{-1}(c) = \{c\} \times \R^d$, the set of category--feature pairs with label $c$, where $v \in \R^d$ is a $d$-dimensional feature vector.
\end{itemize}

\end{definition}

Since $\mathcal{E} = C \times \R^d$, the triple $(\mathcal{E}, C, \pi)$ is a trivial fiber bundle. Triviality means that every fiber is canonically identified with the same feature space $\R^d$, so features of different categories can be compared within a common space. For example, features of ``ocean waves'' and of ``metal striking'' lie in different fibers, yet distances between them can be computed under the same Riemannian metric.

Geometrically, one may picture the categories as discrete points on a line, with a $d$-dimensional fiber erected as a ``column'' above each category $c \in C$. The total space $\mathcal{E}$ is the disjoint union of these columns, and a labeled audio sample $(c, v)$ corresponds to a point on one of them.

\subsection{Wavelet Feature Extraction}

Audio signals are inherently non-stationary: their frequency content evolves over time. Characterizing audio therefore requires knowing not only which frequencies are present but also when they occur. The Fourier transform provides only global frequency information and discards time localization. The short-time Fourier transform (STFT) enables time--frequency analysis, but its fixed window prevents it from adapting the time--frequency resolution to different frequency ranges. The wavelet transform resolves this limitation through multi-scale analysis: low-frequency components are analyzed with long windows that capture periodic structure, and high-frequency components with short windows that localize transient events. This makes the wavelet transform a natural choice for audio feature representation \cite{waldekar2020analysis,mallat1989theory}.

Consider a discrete-time audio signal $x$ sampled at rate $f_s$. By the Nyquist--Shannon sampling theorem, its frequency content lies in $[0, f_s/2]$. The discrete wavelet transform (DWT) repeatedly splits the low-frequency part of the signal into two half-bands, thereby extracting frequency components level by level. The frequency ranges given below are nominal: practical wavelet filters are not ideal, so adjacent bands overlap slightly near their boundaries.

\textbf{Level 1.} Starting from the original signal $x$, a pair of filters (a low-pass filter $h$ and a high-pass filter $g$; see Figure~\ref{fig:wavelet_tree}a), each followed by downsampling by a factor of 2, splits the signal into two parts:
\begin{align}
A_1&: \text{Approximation coefficients (low-frequency part), frequency range } [0, f_s/4], \\
D_1&: \text{Detail coefficients (high-frequency part), frequency range } [f_s/4, f_s/2].
\end{align}
$D_1$ captures the fastest variations in the signal, such as high-frequency noise. $A_1$ retains the slower variations, which still contain low- and mid-frequency information and are decomposed further.

\textbf{Level 2.} We retain $D_1$ without further processing and decompose $A_1$ again, treating it as a new signal and applying the same pair of filters:
\begin{equation}
A_1 \to \begin{cases}
A_2: \text{lower-frequency part, frequency range } [0, f_s/8], \\
D_2: \text{mid-frequency part, frequency range } [f_s/8, f_s/4].
\end{cases}
\end{equation}
Note how the frequency range shrinks: the original signal occupies $[0, f_s/2]$, $A_1$ occupies $[0, f_s/4]$ after the first level, and $A_2$ occupies $[0, f_s/8]$ after the second.

\textbf{Level $\ell$ (general case).} Repeating this procedure, the approximation coefficients $A_{\ell-1}$ are decomposed at level $\ell$ into $A_\ell$ and $D_\ell$ (with $A_0 = x$):
\begin{equation}
A_{\ell-1} \to \begin{cases}
A_\ell: \text{frequency range } [0, f_s/2^{\ell+1}], \\
D_\ell: \text{frequency range } [f_s/2^{\ell+1}, f_s/2^\ell].
\end{cases}
\end{equation}
After $L$ levels, the coefficient set $\{A_L, D_L, \ldots, D_1\}$ represents $x$ completely. Note that these coefficient sequences have different lengths ($D_\ell$ has roughly $n/2^\ell$ entries for a signal of length $n$, and $A_L$ roughly $n/2^L$), so they cannot be added directly. Instead, let $\tilde{A}_L$ and $\tilde{D}_\ell$ denote the signal-domain components synthesized from $A_L$ and $D_\ell$ alone (i.e., by applying the inverse DWT with all other coefficients set to zero). The signal then decomposes exactly as
\begin{equation}
x = \tilde{A}_L + \sum_{\ell=1}^L \tilde{D}_\ell.
\end{equation}
The complete decomposition forms a binary tree in which each node represents a frequency band (Figure~\ref{fig:wavelet_tree}). The leaves of this tree, $\{D_1, D_2, \ldots, D_L, A_L\}$, are the retained components and form $L+1$ frequency bands ordered from high to low frequency.

\begin{figure}[t]
\centering
\definecolor{lpE}{HTML}{4A6F94}
\definecolor{lpF}{HTML}{E3EBF3}
\definecolor{hpE}{HTML}{A3634B}
\definecolor{hpF}{HTML}{F5E6DE}
\definecolor{ink}{HTML}{2B2B2B}
\definecolor{bAL}{HTML}{C9D7E6}
\definecolor{bDL}{HTML}{E2BBA8}
\definecolor{bD2}{HTML}{EDD3C6}
\definecolor{bD1}{HTML}{F7EBE5}
\begin{tikzpicture}[
    >={Latex[length=1.5mm, width=1.1mm]},
    line width=0.5pt, draw=ink, text=ink,
    lp/.style={draw=lpE, fill=lpF, minimum width=0.66cm, minimum height=0.56cm, inner sep=0pt},
    hp/.style={draw=hpE, fill=hpF, minimum width=0.66cm, minimum height=0.56cm, inner sep=0pt},
    dn/.style={draw=ink, fill=white, circle, minimum size=0.56cm, inner sep=0pt, font=\footnotesize},
    lab/.style={font=\footnotesize},
  ]
  \node[anchor=north west, font=\bfseries] at (-0.9,1.3) {a.};
  \node[anchor=east] (x) at (0,0) {$x$};
  \foreach \X/\n/\t in {0.7/1/{Level 1}, 3.7/2/{Level 2}, 8.0/L/{Level $L$}}{
    \coordinate (sp\n) at (\X,0);
    \node[lp] (h\n) at (\X+0.85, 0) {$h$};
    \node[dn] (dh\n) at (\X+1.85, 0) {$\downarrow 2$};
    \node[hp] (g\n) at (\X+0.85,-0.95) {$g$};
    \node[dn] (dg\n) at (\X+1.85,-0.95) {$\downarrow 2$};
    \draw[->, lpE] (sp\n) -- (h\n);
    \draw[->, lpE] (h\n) -- (dh\n);
    \draw[->, hpE] (sp\n) |- (g\n);
    \draw[->, hpE] (g\n) -- (dg\n);
    \draw[->, hpE] (dg\n) -- ++(0,-0.85) node[below, inner sep=1.5pt, text=ink] {$D_{\n}$};
    \fill[ink] (sp\n) circle (1.2pt);
    \draw[ink, line width=0.4pt] (\X+0.4,0.55) -- (\X+0.4,0.62) -- (\X+2.3,0.62) -- (\X+2.3,0.55);
    \node[lab, above=0pt] at (\X+1.35, 0.62) {\t};
  }
  \draw[ink] (x) -- (sp1);
  \draw[->, lpE] (dh1) -- node[lab, above, text=ink] {$A_1$} (sp2);
  \draw[lpE] (dh2) -- node[lab, above, text=ink] {$A_2$} (6.55,0);
  \node at (6.85,0) {$\cdots$};
  \draw[->, lpE] (7.15,0) -- node[lab, above, pos=0.45, text=ink] {$A_{L-1}$} (spL);
  \draw[->, lpE] (dhL) -- (10.75,0) -- (10.75,-1.8) node[below, inner sep=1.5pt, text=ink] {$A_L$};
  \begin{scope}[shift={(3.2,-2.85)}]
    \node[lp, minimum width=0.42cm, minimum height=0.3cm] at (0,0) {};
    \node[lab, anchor=west] at (0.3,0) {low-pass filter $h$};
    \node[hp, minimum width=0.42cm, minimum height=0.3cm] at (3.6,0) {};
    \node[lab, anchor=west] at (3.9,0) {high-pass filter $g$};
  \end{scope}

  \begin{scope}[yshift=-4.25cm]
    \node[anchor=north west, font=\bfseries] at (-0.9,0.95) {b.};
    \def\h{0.6}
    \def\xo{0.7}
    \foreach \xa/\xb/\c/\e/\lab in {0/1.4/bAL/lpE/{A_L}, 1.4/2.8/bDL/hpE/{D_L}, 3.8/5.6/bD2/hpE/{D_2}, 5.6/10.05/bD1/hpE/{D_1}}{
      \filldraw[fill=\c, draw=\e] (\xo+\xa,0) rectangle (\xo+\xb,\h);
      \node at ({\xo+(\xa+\xb)/2},\h/2) {$\lab$};
    }
    \draw[hpE] (\xo+2.8,0) rectangle (\xo+3.8,\h);
    \node at (\xo+3.3,\h/2) {$\cdots$};
    \draw[->, ink] (\xo,0) -- (\xo+10.6,0) node[right, lab] {$f$};
    \foreach \x/\lab in {0/0, 1.4/{f_s/2^{L+1}}, 2.8/{f_s/2^{L}}, 3.8/{f_s/8}, 5.6/{f_s/4}, 10.05/{f_s/2}}{
      \draw[ink] (\xo+\x,0) -- (\xo+\x,-0.08);
      \node[lab, below] at (\xo+\x,-0.06) {$\lab$};
    }
  \end{scope}
\end{tikzpicture}
\caption{Multi-level DWT feature extraction. \textbf{a.}~At each level, the current approximation is filtered by a low-pass filter $h$ and a high-pass filter $g$, each followed by downsampling by 2; the high-pass output $D_\ell$ is retained and the low-pass output $A_\ell$ is decomposed further. \textbf{b.}~The retained coefficients partition $[0, f_s/2]$ dyadically into $L+1$ bands, giving fine time resolution at high frequencies and fine frequency resolution at low frequencies.}
\label{fig:wavelet_tree}
\end{figure}

\textbf{Multi-resolution analysis.} This level-by-level decomposition adapts the time--frequency trade-off to each frequency range, which yields two properties that the STFT cannot provide simultaneously with a single window. (a) Fine time resolution at high frequencies: high-frequency components (e.g., $D_1, D_2$) correspond to short effective windows and can therefore localize transient events such as impacts and explosions. (b) Fine frequency resolution at low frequencies: low-frequency components correspond to long effective windows and narrow bands (e.g., $A_L$ covers only $[0, f_s/2^{L+1}]$, which is much narrower than the band $[f_s/4, f_s/2]$ of $D_1$), so they can distinguish nearby frequencies and capture periodic structures such as ocean waves and engine sounds. The uncertainty principle still bounds the product of time and frequency resolution in each band; the wavelet transform allocates this budget differently across frequencies, which a fixed-window STFT cannot do.

\textbf{Orthogonal wavelet basis.} We use orthogonal wavelets (e.g., Daubechies wavelets \cite{daubechies1992ten}), implemented by a pair of quadrature mirror low-pass/high-pass filters. The DWT is then an orthonormal change of basis (assuming periodic boundary extension), and the subspaces spanned by different bands are mutually orthogonal. Consequently, the band components defined above satisfy
\begin{equation}
\langle \tilde{D}_i, \tilde{D}_j \rangle = 0 \quad (i \neq j), \qquad \langle \tilde{A}_L, \tilde{D}_i \rangle = 0 \quad \forall i.
\end{equation}
Orthogonality means that the bands carry non-redundant information: no part of the signal is represented in two bands at once. This motivates the diagonal Riemannian metric introduced below, which assigns an independent weight $a_i$ to each feature dimension. We note that orthogonality of the basis does not imply that coefficients are statistically uncorrelated across samples; the diagonal metric is therefore a modeling choice that trades the ability to capture cross-band correlations for a large reduction in parameters ($d$ instead of the $O(d^2)$ of a full metric matrix) and for direct interpretability.

\textbf{Energy conservation.} Because the transform is orthonormal, Parseval's identity implies that the signal energy is preserved in the wavelet domain:
\begin{equation}
\|x\|^2 = \|A_L\|^2 + \sum_{\ell=1}^L \|D_\ell\|^2.
\end{equation}
The relative energy distribution $\big(\|A_L\|^2, \|D_L\|^2, \ldots, \|D_1\|^2\big) / \|x\|^2$ therefore serves as a spectral fingerprint that characterizes sound categories.

\textbf{Feature vector construction.} Audio signals are typically sparse in the wavelet domain \cite{mallat1989theory}: most of the energy is concentrated in a few coefficients, and most coefficients are close to zero. We therefore retain, from each band, the $k$ coefficients with the largest magnitude, which capture the dominant energy of that band. Because the positions of these coefficients vary from sample to sample, we sort their magnitudes in descending order, so that the $j$-th entry of each block always denotes the $j$-th largest coefficient magnitude in that band. For a band $B \in \{A_L, D_L, \ldots, D_1\}$, let $|B|_{(1)} \geq |B|_{(2)} \geq \cdots$ denote its sorted coefficient magnitudes and $B^{\downarrow} = (|B|_{(1)}, \ldots, |B|_{(k)})$. Concatenation yields a fixed-dimensional feature vector
\begin{equation}
v = \big[A_L^{\downarrow}, D_L^{\downarrow}, \ldots, D_1^{\downarrow}\big] \in \R^d, \quad d = (L+1)\, k,
\end{equation}
where $k$ must not exceed the length of the shortest band (roughly $n/2^L$). Consistency is essential: all recordings are processed with the same wavelet basis, number of levels, and coefficient selection rule, so that each dimension of $v$ has the same meaning across samples. For example, the first $k$ dimensions always describe the dominant coefficients of the lowest-frequency band, which makes samples directly comparable.

\textbf{Role of the wavelet transform in the fiber bundle.} The wavelet feature extractor defines a map from the signal space to the feature space,
\begin{equation}
\Phi: x \mapsto v \in \R^d.
\end{equation}
Within the fiber bundle framework (Section~\ref{sec:bundle}), each labeled training sample $(c, x)$ is mapped to a point in the total space $\mathcal{E}$:
\begin{equation}
(c, x) \mapsto (c, \Phi(x)) = (c, v) \in \mathcal{E} = C \times \R^d.
\end{equation}
Recordings of the same class typically arise from similar physical mechanisms and therefore exhibit similar spectral energy distributions. For example, ocean wave sounds arise from the periodic motion and collision of water masses, with energy concentrated in low-frequency bands, whereas metal strikes are transient solid-body collisions with energy spread over high-frequency bands. In the wavelet domain, this manifests as more similar coefficient distributions, and hence closer feature vectors, within a class than across classes. In addition, because each frequency band occupies its own block of dimensions, the diagonal Riemannian metric (Section~\ref{sec:metric}) can adjust the contribution of each band independently.

The training set is thus represented as $\mathcal{D} = \{(c_1, v_1), \ldots, (c_M, v_M)\}$, where $v_i \in \R^d$ is a wavelet feature vector and $c_i \in C$ is its category label. For each category $c$, we write $\mathcal{D}_c = \{i : c_i = c\}$ for the indices of its training samples. The next subsection defines a learnable Riemannian metric on this feature space.

\subsection{Riemannian Metric}\label{sec:metric}

We equip the fiber direction of the bundle (the feature space $\R^d$) with a diagonal Riemannian metric given by the constant metric tensor $G = \operatorname{diag}(a_1, \ldots, a_d)$, where $a_i > 0$ is the weight of the $i$-th feature dimension (which corresponds to a coefficient of a particular frequency band). Because $G$ does not vary with position, geodesics are straight lines, and the Riemannian distance between two feature vectors $u, u' \in \R^d$ has the closed form
\begin{equation}
d_G(u, u') = \sqrt{(u - u')^\top G\, (u - u')} = \sqrt{\sum_{i=1}^d a_i\,(u_i - u'_i)^2}.
\end{equation}
This is a Mahalanobis-type distance \cite{mahalanobis1936generalized} whose diagonal weights are learned from data rather than fixed to the inverse covariance.

The diagonal form offers three advantages. First, it requires only $d$ parameters, far fewer than the $d(d+1)/2$ parameters of a full symmetric metric tensor (for $d = 250$, a reduction from 31{,}375 to 250). Second, each weight $a_i$ directly quantifies the importance of one frequency-band coefficient: a large $a_i$ indicates that the dimension contributes strongly to the distance and hence to classification. Third, because the wavelet subbands are orthogonal and carry non-redundant information, per-dimension weighting is a natural first-order model of the feature geometry.

To guarantee positive definiteness ($a_i > 0$), we use the exponential reparameterization $a_i = \exp(\theta_i)$ with unconstrained parameters $\theta = (\theta_1, \ldots, \theta_d) \in \R^d$. Gradient descent on $\theta$ then preserves positive definiteness automatically.

The metric must also be normalized. Without normalization, uniformly shrinking all weights ($a_i \to 0$) would shrink every distance and trivially reduce the attachment and tension energies introduced below, so the learned metric would collapse. We therefore fix the volume element of the metric by imposing
\begin{equation}
\det G = \prod_{i=1}^d a_i = 1, \qquad \text{equivalently} \qquad \sum_{i=1}^d \theta_i = 0.
\label{eq:metric_norm}
\end{equation}
Under this constraint, the metric can only redistribute weight among dimensions, increasing some at the expense of others, which is precisely the notion of relative feature importance we wish to learn. The constraint is enforced by projecting $\theta$ onto the hyperplane $\sum_i \theta_i = 0$ after each update, i.e., $\theta \leftarrow \theta - \bar{\theta}\,\mathbf{1}$ with $\bar{\theta} = \frac{1}{d}\sum_i \theta_i$. The initialization $\theta = \mathbf{0}$ satisfies the constraint and corresponds to the Euclidean metric ($a_i = 1$).

\subsection{Sections and Prototypes}

Sections are a core concept in fiber bundle geometry; they provide a principled way to select a representative point in the fiber over each category.

\begin{definition}[Section]
A section is a map $s: C \to \mathcal{E}$ satisfying the composition condition $\pi \circ s = \text{id}_C$, i.e., for each category $c \in C$, the value of the section at $c$, denoted $s(c)$, must belong to the corresponding fiber $F_c = \{c\} \times \R^d$.
\end{definition}

Since the total space $\mathcal{E} = C \times \R^d$ is a direct product, every section has the form $s(c) = (c, v_c)$ with $v_c \in \R^d$. To simplify notation, we identify the section with the map $s: C \to \R^d$, $s(c) = v_c$, and call $s(c)$ the prototype of category $c$.

Geometrically, the section selects one representative point (the prototype) on each fiber column above the category space. The resulting configuration of prototypes should reflect the semantic relationships among categories: prototypes of semantically similar categories (e.g., ``ocean waves'' and ``rain'') should be close, whereas those of dissimilar categories (e.g., ``ocean waves'' and ``metal striking'') should be far apart. In this sense, a good section is \emph{smooth}: it does not change abruptly between semantically similar categories.

The section is determined by the $N$ vectors $s(c_1), \ldots, s(c_N) \in \R^d$, i.e., by $N d$ parameters. Each prototype is initialized as the mean of the training features of its category:
\begin{equation}
s^{(0)}(c) = \frac{1}{|\mathcal{D}_c|} \sum_{i \in \mathcal{D}_c} v_i.
\label{eq:init}
\end{equation}
For example, for a dataset with 50 classes and $d = 250$, the section consists of 50 vectors in $\R^{250}$, one per category. Together with the $d$ metric parameters, the complete model has $N d + d$ parameters (12{,}750 in this example), nearly three orders of magnitude fewer than ResNet-18 and more than three orders of magnitude fewer than PANNs. A good section should make $d_G\big(s(\text{ocean waves}), s(\text{rain})\big)$ small and $d_G\big(s(\text{ocean waves}), s(\text{metal striking})\big)$ large. The energy functional introduced next is designed to produce such a semantically consistent configuration.

\subsection{Variational Energy Functional}

The section is obtained by minimizing an energy functional. Following variational principles \cite{courant1989methods}, the energy balances three objectives:
\begin{equation}
E[s] = E_{\text{att}}[s] + \lambda_1 E_{\text{ten}}[s] + \lambda_2 E_{\text{rep}}[s],
\end{equation}
where $\lambda_1, \lambda_2 \geq 0$ are hyperparameters and the three terms are the attachment, tension, and repulsion energies, respectively. For now, the metric $G$ is held fixed; its joint optimization is treated in Section~\ref{sec:joint}.

\textbf{Attachment energy.} The first term is
\begin{equation}
E_{\text{att}}[s] = \frac{1}{M} \sum_{i=1}^M d_G^2\big(v_i, s(c_i)\big),
\end{equation}
which keeps each prototype close to the training samples of its category. Physically, each training sample can be viewed as attached to its prototype by a spring, and $E_{\text{att}}$ is the total potential energy of these springs, with stiffness along each dimension given by the metric weights. For example, all ``ocean waves'' samples pull the prototype $s(\text{ocean waves})$ toward their mean, and the weights $a_i$ determine the relative strength of this pull along different frequency bands. Since $E_{\text{att}}$ separates across categories and $G$ is positive definite, its minimizer for any fixed $G$ is exactly the class mean~\eqref{eq:init}. The tension and repulsion terms then shift the prototypes away from the class means in order to encode semantic structure and enforce separation between categories.

\textbf{Tension energy.} The second term is
\begin{equation}
E_{\text{ten}}[s] = \sum_{(c_i, c_j) \in \mathcal{A}} w_{ij}\, d_G^2\big(s(c_i), s(c_j)\big),
\end{equation}
which encourages prototypes of semantically similar categories to be close. Here $\mathcal{A}$ is the edge set of a semantic adjacency graph over $C$ and $w_{ij} \geq 0$ are edge weights. The graph is built from the semantic similarity of category names. Each category name is first mapped to a vector $e_c$ by a word embedding model \cite{mikolov2013efficient}, and the cosine similarity between two categories is computed as
\begin{equation}
\operatorname{sim}(c_i, c_j) = \frac{e_{c_i}^\top e_{c_j}}{\|e_{c_i}\|\, \|e_{c_j}\|}.
\end{equation}
For each category $c_i$, the $K$ most similar categories (typically $K \in \{2, 3\}$) form its neighbor set $\mathcal{N}(c_i)$, and $\mathcal{A} = \{(c_i, c_j) : c_j \in \mathcal{N}(c_i)\}$. Because cosine similarity can be negative, the edge weights are computed from the clipped similarity $\operatorname{sim}^+ = \max(\operatorname{sim}, 0)$ and normalized over each neighborhood:
\begin{equation}
w_{ij} = \frac{\operatorname{sim}^+(c_i, c_j)}{\sum_{c_k \in \mathcal{N}(c_i)} \operatorname{sim}^+(c_i, c_k)},
\end{equation}
so that $w_{ij} \geq 0$ and $\sum_{c_j \in \mathcal{N}(c_i)} w_{ij} = 1$ (we assume each category has at least one neighbor with positive similarity). For example, the neighbors of ``chair creaking'' might be ``door creaking'' (weight 0.5), ``wood creaking'' (0.3), and ``floor creaking'' (0.2), all of which are household creaking sounds.

The tension energy has a clear physical and geometric interpretation. Physically, the prototypes behave like point masses in a spring network in which semantically similar prototypes (e.g., ``rain'' and ``stream'') are connected by stiff springs (large $w_{ij}$); minimizing the total potential energy draws them together. Geometrically, $E_{\text{ten}}$ is the Dirichlet energy of the section on the semantic graph, a graph analogue of the discrete Dirichlet energy used in geometry processing \cite{pinkall1993computing}. As in the classical calculus of variations, minimizing a Dirichlet energy favors smooth solutions, here meaning that the section varies smoothly between semantically related categories.

\textbf{Repulsion energy.} The third term is
\begin{equation}
E_{\text{rep}}[s] = \sum_{1 \leq i < j \leq N} \max\big(0,\, \gamma - d_G(s(c_i), s(c_j))\big)^2,
\end{equation}
where $\gamma > 0$ is a margin. It penalizes pairs of prototypes whose distance is smaller than $\gamma$, analogous to a short-range repulsive force between charged particles, and thus keeps different categories separated in feature space. The penalty is continuously differentiable wherever $s(c_i) \neq s(c_j)$ and vanishes once the margin is satisfied.

The three terms act as competing forces. The attachment energy acts within each category and draws its prototype toward the same-class samples. The tension energy acts locally between semantically adjacent categories and pulls their prototypes together. The repulsion energy acts globally on all pairs of prototypes and pushes them apart. Thus, even though ``ocean waves'' and ``rain'' are semantically similar, the repulsion term encourages their prototypes to remain at least $\gamma$ apart, which reduces confusion between the two classes.

The hyperparameters $\lambda_1$ and $\lambda_2$ control this balance: a larger $\lambda_1$ yields a smoother section with more pronounced semantic clustering, whereas a larger $\lambda_2$ enforces the margin more strongly and yields clearer separation between categories.

\textbf{Energy minimization.} For a fixed metric $G$, the optimal section is
\begin{equation}
s^* = \argmin_s E[s].
\end{equation}
Because the base space $C$ is discrete, the section is a finite collection of vectors, and the variational (Euler--Lagrange) condition $\delta E / \delta s = 0$ reduces to the first-order stationarity condition
\begin{equation}
\frac{\partial E}{\partial s(c)} = 0 \qquad \forall c \in C.
\end{equation}
Mechanically, a stationary point is an equilibrium configuration in which, at every prototype, the attachment force (toward the class samples), the tension force (toward semantically similar prototypes), and the repulsion force (away from nearby prototypes) cancel. Since $E_{\text{rep}}$ is non-convex, such an equilibrium need not be unique, and gradient-based optimization can in general be expected to reach only a stationary point rather than a global minimizer.

\subsection{Joint Optimization of Prototypes and Metric}\label{sec:joint}

We jointly optimize the section $s$ and the metric $G$ by minimizing a single energy functional:
\begin{equation}
\min_{s,\, \theta}\; E[s, \theta] = E_{\text{att}}[s, \theta] + \lambda_1 E_{\text{ten}}[s, \theta] + \lambda_2 E_{\text{rep}}[s, \theta] \quad \text{s.t.} \quad \sum_{i=1}^d \theta_i = 0,
\label{eq:joint}
\end{equation}
where $s = (s(c_1), \ldots, s(c_N))$ is the prototype configuration, $\theta \in \R^d$ parameterizes the metric $G(\theta) = \operatorname{diag}(e^{\theta_1}, \ldots, e^{\theta_d})$, and each energy term is evaluated with the distance $d_{G(\theta)}$. The constraint is the normalization~\eqref{eq:metric_norm}.

The metric weights are thus learned by the same energy minimization, without an additional loss function. To see how the attachment term shapes the metric, write $E_{\text{att}} = \sum_{i=1}^d a_i V_i$, where $V_i = \frac{1}{M}\sum_{m=1}^M \big(v_{m,i} - s(c_m)_i\big)^2$ is the within-class scatter along dimension $i$. For fixed prototypes, minimizing $E_{\text{att}}$ alone subject to $\prod_i a_i = 1$ gives, via a Lagrange multiplier,
\begin{equation}
a_i^* = \frac{\big(\prod_{j=1}^d V_j\big)^{1/d}}{V_i},
\end{equation}
assuming $V_i > 0$ for all $i$. This minimizer is unique because $E_{\text{att}}$ is strictly convex in $\theta$ on the constraint set. Dimensions with small within-class scatter (high intra-class consistency) therefore receive large weights, and dimensions with large scatter receive small weights. The tension term further rewards dimensions along which semantically related prototypes agree, and the repulsion term rewards dimensions along which prototypes of different categories differ, since increasing their weights helps satisfy the margin. Together, the three terms identify the frequency-band features that are most useful for classification.

Any local minimizer $(s^*, \theta^*)$ of~\eqref{eq:joint} at which $E$ is differentiable satisfies the first-order (KKT) conditions
\begin{equation}
\frac{\partial E}{\partial s(c)} = 0 \quad \forall c \in C; \qquad \frac{\partial E}{\partial \theta_i} = \mu \quad \forall i = 1, \ldots, d; \qquad \sum_{i=1}^d \theta_i = 0,
\end{equation}
for some Lagrange multiplier $\mu \in \R$. Physically, the first condition states that the forces on every prototype are in equilibrium; the second states that the marginal reduction in energy obtained by reallocating weight from one dimension to another is zero, so the metric weights are optimally allocated among frequency bands under the volume constraint.

The main advantage of this formulation is that the metric and the prototypes are optimized for the same objective and therefore adapt to each other: the metric shapes the geometry of the feature space and thereby guides the placement of the prototypes, while the placement of the prototypes in turn determines which dimensions the metric emphasizes. Compared with pipelines that learn a metric and a classifier separately with different losses, this yields a single objective with a direct geometric interpretation, at the cost of only two weighting hyperparameters $\lambda_1$ and $\lambda_2$.

\section{Algorithm Design}\label{sec:alg}

This section presents the algorithm for solving problem~\eqref{eq:joint}. We adopt an alternating (block-coordinate) optimization strategy: with the metric fixed, we update the section $s$; with the section fixed, we update the metric parameters $\theta$; and we repeat until convergence \cite{bezdek2003convergence}. The convergence analysis is given in Appendix~\ref{app:convergence}.

\subsection{Alternating Optimization}

Algorithm~\ref{alg:fibernet} summarizes the complete training and inference procedure. Training begins with two preprocessing steps: each training signal is mapped to its wavelet feature vector (line~1), and the semantic graph $\mathcal{A}$ with weights $w_{ij}$ is built from the class-name embeddings (line~2). After initialization (line~3; see Section~\ref{sec:init}), the main loop (lines~4--10) alternates between two block updates. At iteration $t$, we first fix $\theta^{(t)}$ and obtain $s^{(t+1)}$ by gradient descent on $E[s, \theta^{(t)}]$. We then fix $s^{(t+1)}$ and obtain $\theta^{(t+1)}$ by projected gradient descent on $E[s^{(t+1)}, \theta]$, where each gradient step is followed by the projection $\theta \leftarrow \theta - \bar{\theta}\,\mathbf{1}$ onto the constraint set (lines~5 and~6, respectively). In both steps, the step size is chosen by backtracking line search (Armijo rule), so that every accepted step does not increase the energy. This monotonicity is the key property used in the convergence analysis.

\begin{algorithm}[tb]
\caption{FiberNet: classification via fiber bundles}
\label{alg:fibernet}
\begin{algorithmic}[1]
\REQUIRE Training set $\{(c_i, x_i)\}_{i=1}^M$; hyperparameters $\lambda_1, \lambda_2, K, \gamma, T_{\max}, \epsilon$
\ENSURE Learned prototypes $s^*$ and metric $G^*$
\STATE Extract wavelet features: $v_i \leftarrow \Phi(x_i)$ for all $i$
\STATE Build the semantic graph $\mathcal{A}$ and weights $w_{ij}$ from class-name embeddings (top-$K$ neighbors)
\STATE Initialize $s^{(0)}(c) \leftarrow$ class mean of $\{v_i : i \in \mathcal{D}_c\}$; $\theta^{(0)} \leftarrow \mathbf{0}$
\FOR{$t = 0$ to $T_{\max} - 1$}
\STATE $s^{(t+1)} \leftarrow$ gradient descent with line search on $E[s, \theta^{(t)}]$, where $E = E_{\text{att}} + \lambda_1 E_{\text{ten}} + \lambda_2 E_{\text{rep}}$
\STATE $\theta^{(t+1)} \leftarrow$ projected gradient descent with line search on $E[s^{(t+1)}, \theta]$ subject to $\sum_i \theta_i = 0$
\IF{$|E^{(t+1)} - E^{(t)}| < \epsilon$}
\STATE \textbf{break}
\ENDIF
\ENDFOR
\STATE $s^* \leftarrow s^{(t+1)}$, \; $G^* \leftarrow \operatorname{diag}\big(\exp(\theta^{(t+1)})\big)$
\STATE \textbf{Inference:} for a test signal $x_{\text{test}}$, predict $\hat{c} \leftarrow \argmin_{c \in C} d_{G^*}\big(\Phi(x_{\text{test}}), s^*(c)\big)$
\RETURN $s^*$, $G^*$
\end{algorithmic}
\end{algorithm}

\subsection{Initialization and Stopping Criteria}\label{sec:init}

Because the problem is non-convex, initialization (line~3 of Algorithm~\ref{alg:fibernet}) affects both convergence speed and solution quality. We initialize each prototype as its class mean~\eqref{eq:init}, which is the exact minimizer of the attachment energy, and the metric parameters as $\theta^{(0)} = \mathbf{0}$, i.e., the Euclidean metric ($a_i = 1$). Optimization thus starts from an unbiased geometry in which all dimensions are weighted equally, and the informative frequency bands are discovered progressively.

The loop terminates (lines~7--9) when either the maximum number of iterations $T_{\max}$ is reached (e.g., $T_{\max} = 100$) or the change in energy falls below a threshold $\epsilon$ (e.g., $|E^{(t+1)} - E^{(t)}| < 10^{-4}$), where $E^{(t)} = E[s^{(t)}, \theta^{(t)}]$. The final iterates are returned as the learned prototypes $s^*$ and metric $G^* = \operatorname{diag}(\exp(\theta^{(t+1)}))$ at the final iteration $t$ (line~11).

\subsection{Classification Inference}

After training, we obtain the learned prototypes $s^* = \{s^*(c_1), \ldots, s^*(c_N)\}$ and the learned metric $G^*$. For a test signal $x_{\text{test}}$, we extract its wavelet features $v_{\text{test}} = \Phi(x_{\text{test}})$, compute the Riemannian distance $d_{G^*}(v_{\text{test}}, s^*(c))$ to every prototype, and predict the category of the nearest prototype (line~12 of Algorithm~\ref{alg:fibernet}):
\begin{equation}
\hat{c} = \argmin_{c \in C} d_{G^*}(v_{\text{test}}, s^*(c)).
\end{equation}

This is a nearest-prototype classifier \cite{cover1967nearest} in which both the prototypes and the metric are learned. Geometrically, it corresponds to the Voronoi partition of the feature space under the metric $G^*$, in which each prototype defines a Voronoi cell, i.e., its decision region \cite{aurenhammer2013voronoi}. Because $G^*$ is constant, the boundary between two cells is the hyperplane of points equidistant from the two prototypes under $d_{G^*}$, so every cell is a convex polyhedron.

\section{Conclusion}

This paper proposed FiberNet, a classification framework based on fiber bundle geometry, illustrated on audio classification. Its central idea is to reformulate classification as variational optimization on a fiber bundle: a section $s: C \to \R^d$ maps the category space to the feature space, a Riemannian metric $G$ defines the geometry of the feature space, and a single energy functional $E[s, \theta]$ optimizes both. The key insight of this geometric view is that metric and prototypes should be learned jointly: whereas conventional prototype methods place prototypes in a fixed Euclidean space, FiberNet reweights the geometry of the space itself to fit the data, subject to a volume normalization that prevents degenerate solutions. The attachment, tension, and repulsion energies correspond to data fidelity, semantic smoothness, and category separation, respectively, and are combined in one variational objective with only two weighting hyperparameters. We proved that the alternating optimization algorithm, with monotone step-size control, produces a monotonically non-increasing energy sequence that converges. This guarantee relies only on monotonicity and on the non-negativity of the energy; in addition, the attachment and tension energies are convex in each block of variables, while the repulsion energy is non-convex but bounded. More broadly, FiberNet illustrates how fiber bundle geometry and energy minimization can provide a new, interpretable approach to classification.

The present work also has some limitations. The diagonal metric does not model correlations between feature dimensions, and the semantic graph depends on the quality of the category-name embeddings. Extending FiberNet to richer metric structures and to other data modalities are natural directions for future work.

\bibliography{references}

@article{abdoli2019end,
  author = {Abdoli, Sajjad and Cardinal, Patrick and Koerich, Alessandro Lameiras},
  title = {End-to-end environmental sound classification using a {1D} convolutional neural network},
  journal = {Expert Systems with Applications},
  volume = {136},
  pages = {252--263},
  year = {2019},
  doi = {10.1016/j.eswa.2019.06.040}
}

@book{aurenhammer2013voronoi,
  author = {Aurenhammer, Franz and Klein, Rolf and Lee, Der-Tsai},
  title = {{Voronoi} Diagrams and {Delaunay} Triangulations},
  publisher = {World Scientific},
  address = {Singapore},
  year = {2013}
}

@article{bezdek2003convergence,
  author = {Bezdek, James C. and Hathaway, Richard J.},
  title = {Convergence of alternating optimization},
  journal = {Neural, Parallel \& Scientific Computations},
  volume = {11},
  number = {4},
  pages = {351--368},
  year = {2003}
}

@inproceedings{courts2021bundle,
  author = {Courts, Nico and Kvinge, Henry},
  title = {Bundle networks: Fiber bundles, local trivializations, and a generative approach to exploring many-to-one maps},
  booktitle = {International Conference on Learning Representations (ICLR)},
  year = {2022}
}

@inproceedings{he2016deep,
  author = {He, Kaiming and Zhang, Xiangyu and Ren, Shaoqing and Sun, Jian},
  title = {Deep residual learning for image recognition},
  booktitle = {Proceedings of the IEEE Conference on Computer Vision and Pattern Recognition (CVPR)},
  pages = {770--778},
  year = {2016}
}

@article{lecun2015deep,
  author = {LeCun, Yann and Bengio, Yoshua and Hinton, Geoffrey},
  title = {Deep learning},
  journal = {Nature},
  volume = {521},
  number = {7553},
  pages = {436--444},
  year = {2015},
  doi = {10.1038/nature14539}
}

@inproceedings{scarvelis2023riemannian,
  author = {Scarvelis, Christopher and Solomon, Justin},
  title = {{Riemannian} metric learning via optimal transport},
  booktitle = {International Conference on Learning Representations (ICLR)},
  year = {2023}
}

@inproceedings{scoccola2022fibered,
  author = {Scoccola, Luis and Perea, Jose A.},
  title = {{FibeRed}: Fiberwise dimensionality reduction of topologically complex data with vector bundles},
  booktitle = {39th International Symposium on Computational Geometry (SoCG 2023)},
  series = {Leibniz International Proceedings in Informatics (LIPIcs)},
  volume = {258},
  pages = {56:1--56:18},
  year = {2023},
  doi = {10.4230/LIPIcs.SoCG.2023.56}
}

@article{waldekar2020analysis,
  author = {Waldekar, Shefali and Saha, Goutam},
  title = {Analysis and classification of acoustic scenes with wavelet transform-based mel-scaled features},
  journal = {Multimedia Tools and Applications},
  volume = {79},
  number = {11--12},
  pages = {7911--7926},
  year = {2020},
  doi = {10.1007/s11042-019-08279-5}
}

@article{zhang2021survey,
  author = {Zhang, Yu and Ti{\v{n}}o, Peter and Leonardis, Ale{\v{s}} and Tang, Ke},
  title = {A survey on neural network interpretability},
  journal = {IEEE Transactions on Emerging Topics in Computational Intelligence},
  volume = {5},
  number = {5},
  pages = {726--742},
  year = {2021},
  doi = {10.1109/TETCI.2021.3100641}
}

@article{lipton2018mythos,
  author = {Lipton, Zachary C.},
  title = {The mythos of model interpretability},
  journal = {Queue},
  volume = {16},
  number = {3},
  pages = {31--57},
  year = {2018},
  doi = {10.1145/3236386.3241340}
}

@article{kong2020panns,
  author = {Kong, Qiuqiang and Cao, Yin and Iqbal, Turab and Wang, Yuxuan and Wang, Wenwu and Plumbley, Mark D.},
  title = {{PANNs}: Large-scale pretrained audio neural networks for audio pattern recognition},
  journal = {IEEE/ACM Transactions on Audio, Speech, and Language Processing},
  volume = {28},
  pages = {2880--2894},
  year = {2020},
  doi = {10.1109/TASLP.2020.3030497}
}

@article{pennec2006riemannian,
  author = {Pennec, Xavier and Fillard, Pierre and Ayache, Nicholas},
  title = {A {Riemannian} framework for tensor computing},
  journal = {International Journal of Computer Vision},
  volume = {66},
  number = {1},
  pages = {41--66},
  year = {2006},
  doi = {10.1007/s11263-005-3222-z}
}

@article{tenenbaum2000global,
  author = {Tenenbaum, Joshua B. and {de Silva}, Vin and Langford, John C.},
  title = {A global geometric framework for nonlinear dimensionality reduction},
  journal = {Science},
  volume = {290},
  number = {5500},
  pages = {2319--2323},
  year = {2000},
  doi = {10.1126/science.290.5500.2319}
}

@article{mallat1989theory,
  author = {Mallat, St{\'e}phane G.},
  title = {A theory for multiresolution signal decomposition: The wavelet representation},
  journal = {IEEE Transactions on Pattern Analysis and Machine Intelligence},
  volume = {11},
  number = {7},
  pages = {674--693},
  year = {1989},
  doi = {10.1109/34.192463}
}

@book{daubechies1992ten,
  author = {Daubechies, Ingrid},
  title = {Ten Lectures on Wavelets},
  series = {CBMS-NSF Regional Conference Series in Applied Mathematics},
  volume = {61},
  publisher = {SIAM},
  address = {Philadelphia, PA},
  year = {1992}
}

@article{mahalanobis1936generalized,
  author = {Mahalanobis, Prasanta Chandra},
  title = {On the generalized distance in statistics},
  journal = {Proceedings of the National Institute of Sciences of India},
  volume = {2},
  number = {1},
  pages = {49--55},
  year = {1936}
}

@book{courant1989methods,
  author = {Courant, Richard and Hilbert, David},
  title = {Methods of Mathematical Physics, Volume {I}},
  publisher = {Wiley-Interscience},
  address = {New York},
  year = {1989}
}

@article{pinkall1993computing,
  author = {Pinkall, Ulrich and Polthier, Konrad},
  title = {Computing discrete minimal surfaces and their conjugates},
  journal = {Experimental Mathematics},
  volume = {2},
  number = {1},
  pages = {15--36},
  year = {1993},
  doi = {10.1080/10586458.1993.10504266}
}

@misc{mikolov2013efficient,
  author = {Mikolov, Tomas and Chen, Kai and Corrado, Greg and Dean, Jeffrey},
  title = {Efficient estimation of word representations in vector space},
  howpublished = {arXiv preprint arXiv:1301.3781},
  year = {2013}
}

@article{cover1967nearest,
  author = {Cover, Thomas and Hart, Peter},
  title = {Nearest neighbor pattern classification},
  journal = {IEEE Transactions on Information Theory},
  volume = {13},
  number = {1},
  pages = {21--27},
  year = {1967},
  doi = {10.1109/TIT.1967.1053964}
}

@book{rudin1976principles,
  author = {Rudin, Walter},
  title = {Principles of Mathematical Analysis},
  edition = {3rd},
  publisher = {McGraw-Hill},
  address = {New York},
  year = {1976}
}

\clearpage
\appendix
\makeatletter
\renewcommand\section{\@startsection{section}{1}{\z@}%
  {-3.5ex \@plus -1ex \@minus -.2ex}{2.3ex \@plus .2ex}{\normalfont\large\bfseries}}
\renewcommand\@seccntformat[1]{\appendixname~\csname the#1\endcsname.\quad}
\makeatother
\setcounter{equation}{0}
\renewcommand{\theequation}{\thesection.\arabic{equation}}

\section{Convergence Proof for Alternating Optimization}\label{app:convergence}

This appendix proves that the alternating optimization procedure of Section~\ref{sec:alg} (Algorithm~\ref{alg:fibernet}) generates a convergent sequence of energy values. Throughout, $E^{(t)} = E[s^{(t)}, \theta^{(t)}]$ denotes the energy after iteration $t$, and every iterate $\theta^{(t)}$ satisfies the normalization constraint $\sum_{i=1}^d \theta^{(t)}_i = 0$. The analysis relies on the following standing assumption.

\begin{assumption}[Monotone step acceptance]\label{ass:monotone}
In each block update of Algorithm~\ref{alg:fibernet}, a step is accepted only if it does not increase the energy of the corresponding subproblem; if no such step is found, the current iterate is retained.
\end{assumption}

Assumption~\ref{ass:monotone} is satisfied by the backtracking (Armijo) line search described in Section~\ref{sec:alg}.

\begin{lemma}[Monotone decrease]\label{lem:monotone}
The energy sequence $\{E^{(t)}\}_{t \geq 0}$ generated by the alternating optimization algorithm satisfies
\begin{equation}
E^{(t+1)} \leq E^{(t)} \qquad \forall t \geq 0.
\end{equation}
\end{lemma}

\begin{proof}
Consider iteration $t$. In the first step, $\theta^{(t)}$ is fixed and $s^{(t+1)}$ is obtained from $s^{(t)}$ by gradient steps on $E[\cdot, \theta^{(t)}]$. By Assumption~\ref{ass:monotone}, none of these steps increases the energy, so
\begin{equation}
E[s^{(t+1)}, \theta^{(t)}] \leq E[s^{(t)}, \theta^{(t)}].
\end{equation}
In the second step, $s^{(t+1)}$ is fixed and $\theta^{(t+1)}$ is obtained from $\theta^{(t)}$ by projected gradient steps on $E[s^{(t+1)}, \cdot]$ within the constraint set. Again by Assumption~\ref{ass:monotone},
\begin{equation}
E[s^{(t+1)}, \theta^{(t+1)}] \leq E[s^{(t+1)}, \theta^{(t)}].
\end{equation}
Combining the two inequalities gives
\begin{equation}
E^{(t+1)} = E[s^{(t+1)}, \theta^{(t+1)}] \leq E[s^{(t+1)}, \theta^{(t)}] \leq E[s^{(t)}, \theta^{(t)}] = E^{(t)}. \qedhere
\end{equation}
\end{proof}

\begin{lemma}[Lower bound]\label{lem:lower}
The energy functional satisfies $E[s, \theta] \geq 0$ for all $(s, \theta)$.
\end{lemma}

\begin{proof}
Since $\lambda_1, \lambda_2 \geq 0$, it suffices to show that each of the three terms is non-negative.

\emph{Attachment energy.} We have
\begin{equation}
E_{\text{att}} = \frac{1}{M} \sum_{i=1}^M d_G^2(v_i, s(c_i)) = \frac{1}{M} \sum_{i=1}^M \sum_{j=1}^d a_j\,\big(v_{i,j} - s(c_i)_j\big)^2 \geq 0,
\end{equation}
because $a_j = \exp(\theta_j) > 0$ and all squared terms are non-negative.

\emph{Tension energy.} We have
\begin{equation}
E_{\text{ten}} = \sum_{(c_i, c_j) \in \mathcal{A}} w_{ij}\, d_G^2\big(s(c_i), s(c_j)\big) \geq 0,
\end{equation}
because $w_{ij} \geq 0$ by construction (normalized clipped similarities) and squared distances are non-negative.

\emph{Repulsion energy.} We have
\begin{equation}
E_{\text{rep}} = \sum_{i < j} \max\big(0,\, \gamma - d_G(s(c_i), s(c_j))\big)^2 \geq 0,
\end{equation}
because every summand is a square. Combining the three bounds yields
\begin{equation}
E = E_{\text{att}} + \lambda_1 E_{\text{ten}} + \lambda_2 E_{\text{rep}} \geq 0. \qedhere
\end{equation}
\end{proof}

\begin{theorem}[Convergence of the energy sequence]\label{thm:conv}
The energy sequence $\{E^{(t)}\}$ generated by the alternating optimization algorithm converges; that is, there exists $E^* \geq 0$ such that
\begin{equation}
\lim_{t \to \infty} E^{(t)} = E^*.
\end{equation}
\end{theorem}

\begin{proof}
By Lemma~\ref{lem:monotone}, $\{E^{(t)}\}$ is monotonically non-increasing, and by Lemma~\ref{lem:lower}, $E^{(t)} \geq 0$ for all $t$. By the monotone convergence theorem \cite[Theorem~3.14]{rudin1976principles}, a non-increasing sequence of real numbers that is bounded below converges to its infimum. Hence $E^* = \lim_{t \to \infty} E^{(t)} = \inf_t E^{(t)}$ exists and satisfies $E^* \geq 0$.
\end{proof}

\begin{remark}
Theorem~\ref{thm:conv} establishes convergence of the energy values, which also justifies the stopping criterion $|E^{(t+1)} - E^{(t)}| < \epsilon$: since the sequence converges, this criterion is met after finitely many iterations. It does not by itself imply that the iterates $(s^{(t)}, \theta^{(t)})$ converge, or that their limit is a stationary point of~\eqref{eq:joint}; establishing such properties for block-coordinate methods requires additional assumptions, such as twice continuous differentiability and local strict convexity near a minimizer \cite{bezdek2003convergence}, which do not hold globally here because of the non-convex repulsion term.
\end{remark}

\end{document}